\documentclass[preprint,12pt]{elsarticle}

\usepackage{amssymb}
\usepackage{soulutf8,color}

\journal{Journal of Biomedical Informatics}

\begin{document}

\begin{frontmatter}



\title{The RareDis corpus: a corpus annotated with rare diseases, their signs and symptoms}


\author[inst1]{Claudia Martínez-deMiguel}

\affiliation[inst1]{organization={Tissue Engineering and Regenerative Medicine group, Department of Bioengineering, Universidad Carlos III de Madrid},
            addressline={Avenidad de la Universidad, 30}, 
            city={Leganés},
            postcode={28911}, 
            state={Madrid},
            country={Spain}}

\author[inst2]{Isabel Segura-Bedmar}
\author[inst1,inst3,inst4]{Esteban Chacón-Solano}
\author[inst1,inst3,inst4,inst5]{Sara Guerrero-Aspizua}

\affiliation[inst2]{organization={Human Language and Accesibility Technologies, Computer Science Department},
            addressline={Avenidad de la Universidad, 30}, 
            city={Leganés},
            postcode={28911}, 
            state={Madrid},
            country={Spain}}
            
\affiliation[inst3]{organization={Hospital Fundación Jiménez Díaz e Instituto de Investigación, FJD},
            addressline={Av. de los Reyes Católicos, 2}, 
            city={Madrid},
            postcode={28040}, 
            state={Madrid},
            country={Spain}}

\affiliation[inst4]{organization={Epithelial Biomedicine Division, CIEMAT},
            city={Madrid},
            postcode={28040}, 
            state={Madrid},
            country={Spain}}
\affiliation[inst5]{organization={Centre for Biomedical Network Research on Rare Diseases (CIBERER)},
            addressline={C/Monforte de Lemos 3-5}, 
            city={Madrid},
            postcode={28029}, 
            state={Madrid},
            country={Spain}}

\begin{abstract}
Rare diseases affect a small number of people compared to the general population. However, more than 6,000 different rare diseases exist and, in total, they affect more than 300 million people worldwide. Rare diseases share as part of their main problem, the delay in diagnosis and the sparse information available for  researchers, clinicians, and patients. Finding a diagnostic can be a very long and frustrating experience for patients and their families. The average diagnostic delay is between 6-8 years. Many of these diseases result in different manifestations among patients, which hampers even more their detection and the correct treatment choice. Therefore, there is an urgent need to increase the scientific and medical knowledge about rare diseases. Natural Language Processing (NLP) can help to extract relevant information about rare diseases to facilitate their diagnosis and treatments, but most NLP techniques require manually annotated corpora. Therefore, our goal is to create a gold standard corpus annotated with rare diseases and their clinical manifestations. It could be used to train and test NLP approaches and the information extracted through NLP could enrich the knowledge of rare diseases, and thereby, help to reduce the diagnostic delay and improve the treatment of rare diseases.
The paper describes the selection of 1,041 texts to be included in the corpus, the annotation process and the annotation guidelines. The entities (\emph{disease, rare disease, symptom, sign and anaphor}) and the relationships (\emph{produces, is a, is acron, is synon, increases risk of, anaphora}) were annotated. 
The RareDis corpus contains more than 5,000 rare diseases and almost 6,000 clinical manifestations are annotated. Moreover, the Inter Annotator Agreement evaluation shows a relatively high agreement (F1-measure equal to 83.5\% under exact match criteria for the entities and equal to 81.3\% for the relations).
Based on these results, this corpus is of high quality, supposing a significant step for the field since there is a scarcity of available corpus annotated with rare diseases. This could open the door to further NLP applications, which would facilitate the diagnosis and treatment of these rare diseases and, therefore, would improve dramatically the quality of life of these patients. 
\end{abstract}

\end{frontmatter}


\section{Introduction}
\label{sec:intro}

Rare diseases affect a small number of people compared to the general population. However, more than 6,000 different rare diseases exist and they affect more than 300 million people worldwide. They are also known as orphan diseases because they are so rare that the development of new therapeutics would not be profitable to produce without government assistance. Indeed, approximately 95\% of rare diseases do not have any treatment and there are roughly only a hundred drugs for these pathologies \cite{klimova2017global}.

Rare diseases share as part of their main problem, the delay in diagnosis and the sparse information available for  researchers, clinicians, and patients. Finding a diagnosis can be a very long and frustrating experience for patients and their families. The average diagnostic delay is around seven years  \citep{ggenes}. Many of these diseases result in different manifestations among patients with the same disease, which hampers even more their detection and the correct treatment choice. Therefore, there is an urgent need to increase the scientific and medical knowledge about rare diseases  \citep{schaefer2020use}. 

Most of the knowledge about rare diseases is encoded in structured sources such as databases and ontologies, but also in written texts such as research articles, clinical cases, clinical trials, drug safety reports, health agency newsletters, as well as information from social media, websites, and health forums  \citep{irdirc}. Thus, having an accurate and complete picture of rare diseases and their signs and symptoms is a very challenging task for researchers and healthcare professionals. 

In the era of Big Data, Natural Language Processing (NLP) has become one of the most relevant research areas to process and analyze large volumes of unstructured information available in any domain or field of knowledge. A clear example is the health domain, where there are multiple sources of textual information. The transformation of the textual information to a structured format can facilitate the access and analysis of the knowledge contained within these multiple sources. Therefore, NLP could help us to extract relevant information about rare diseases thus improving accuracy in their diagnosis and treatment choices. However, most NLP techniques (especially those based on machine learning algorithms)  depend on the existence of large collections of annotated texts that can be used to train and test these techniques. With this purpose, our goal is to create a gold standard corpus annotated with rare diseases, their signs and symptoms. Then, this corpus could be used to train and test different approaches to automatically extract relevant information about rare diseases. The information extracted by NLP approaches could enrich the knowledge of rare diseases, helping to reduce the diagnostic delay and improve the treatment of these diseases.

In the last decade, several competitions such as BioCreative  \citep{hirschman2005overview}, i2b2  \citep{uzuner20112010}, BioNLP shared tasks  \citep{kim2012genia}  and DDIExtraction  \citep{segura2013semeval,segura2014lessons} have contributed significantly to the advance of research in NLP techniques as applied to the domains of biology and biomedicine. As a result, many systems and tools (MetaMap   \citep{aronson2001effective}, cTakes   \citep{savova2010mayo}, MedLEE   \citep{sevenster2012automatically}) have been developed for the recognition of entities and extraction of relationships of these domains. 

These systems require large annotated corpora. There are some corpora including annotations of disease mentions such as the NCBI disease corpus   \citep{dougan2014ncbi}, the EU-ADR corpus   \citep{van2012eu}, the ADE corpus  \citep{gurulingappa2012development}, or the PrevComp corpus \citep{alnazzawi2021building}. 

In the rare diseases domain, very few efforts have been performed \citep{chen2017opportunities,metivier2015automatic,laburu2018can,fabregat2018deep}, due probably to the scarcity of reliable and valid annotated corpora for the training and testing of supervised approaches. 

To the best of our knowledge, before our corpus, there is just a previous one   \citep{fabregat2018deep}, the RDD corpus, annotated with rare diseases. It consists of 1,000 abstracts annotated with rare diseases and disabilities (impairments, activity limitations, etc). The abstracts were initially selected from a list of the Orphanet database  \citep{orphanet1}. 
The annotators (computer science scientists) using the BRAT annotation tool  \citep{stenetorp2012brat}  manually annotated rare diseases and disabilities, including negation and speculation expressions related with a disability. A total of 578 rare diseases and 3,678 disabilities were annotated. The agreement, which was measured as the percentage of coincidences, reaches 87\% for disabilities. The authors did not provide information about the agreement for rare diseases.

In this paper, we present the RareDis corpus, a corpus annotated with rare diseases, their signs and their symptoms. 
We provide a comprehensive description of the annotation process, the annotation guidelines, and the characteristics of the RareDis corpus. The corpus and its guidelines are publicly available for the research community: https://\\github.com/isegura/NLP4RARE-CM-UC3M.
The creation of this kind of corpora can attract the attention of the NLP community to the domain of rare diseases. The use of NLP technology can make a tremendous difference in better understanding rare diseases, helping physicians in their clinical practice and hopefully improving the quality of life of patients, in a the near future.



\section{Methods}
\label{sec:methods}
\subsection{Corpus construction}
\label{subsec:construction}
The RareDis corpus consists of texts taken from the rare disease database \citep{raredb}, created and maintained by the National Organisation for Rare Diseases (NORD). This database contains detailed information about more than 1,200 rare diseases. For each rare disease, the database provides a text organised in the following sections: general discussion, signs and symptoms, causes, affected populations, related disorders, diagnosis, standard therapies, investigational therapies, NORD member organisations and other organisations. We used the seven first sections of each text. To download all the information related to rare diseases, we performed web scraping. This is the process of extracting data from a website by using an automated program. To do this, we developed a Python script based on the use of the Beautiful Soup library \citep{beautisoup}, obtaining a total of 1,041 English texts. 

In order to reduce the heavy workload of manual annotation and accelerate the annotation process, we developed a dictionary-based approach to automatically annotate the mentions of diseases, rare diseases and symptoms in our texts. This method was implemented with Spacy \citep{spacy} and used the following dictionaries:
\begin{itemize}
    \item Disease Ontology (DOID) \citep{schriml2019human} is a standardised ontology for human diseases, which was created by the University of Maryland School of Medicine (Institute for Genome Sciences). It contains 9,871 disease terms and semantically integrates vocabulary from MeSH, \citep{mesh1} International Statistical Classification of Diseases and Related Health Problems (ICD), \citep{icd11clas} NCI thesaurus, \citep{nci2021thesaurus} SNOMED, \citep{snomed2021} and OMIM database \citep{omim}.
    
    \item Orphan Rare Disease Ontology (ORDO) \citep{orphanet1}, developed by Orphanet and the European Bioinformatics Institute (EBI), contains a classification of rare diseases, gene-disease relationships and epidemiological data, as well as mappings to other terminological resources (such as MeSH, OMIM, UMLS \citep{umls}, ICD, \citep{icd11clas} MedDRA \citep{meddra}, UniProtKB \citep{unitprot}, HGNC \citep{hgnc}, ensembl \citep{ensembl}, Reactome \citep{reactome}, IUPHAR \citep{iuphar}). ORDO contains 14,501 classes.
    
    \item Symptom Ontology (SYMP) \citep{symp} contains more than 1,164 terms related to signs, symptoms and diseases. 
\end{itemize}

The automatic pre-annotation identified a total of 3,003 diseases, 2,542 rare diseases and 1,560 symptoms. The pre-annotated texts are the starting point of the annotation process performed by our annotators.


\subsection{Annotation Process}

\begin{figure}
  \centering

    \includegraphics[scale=0.20]{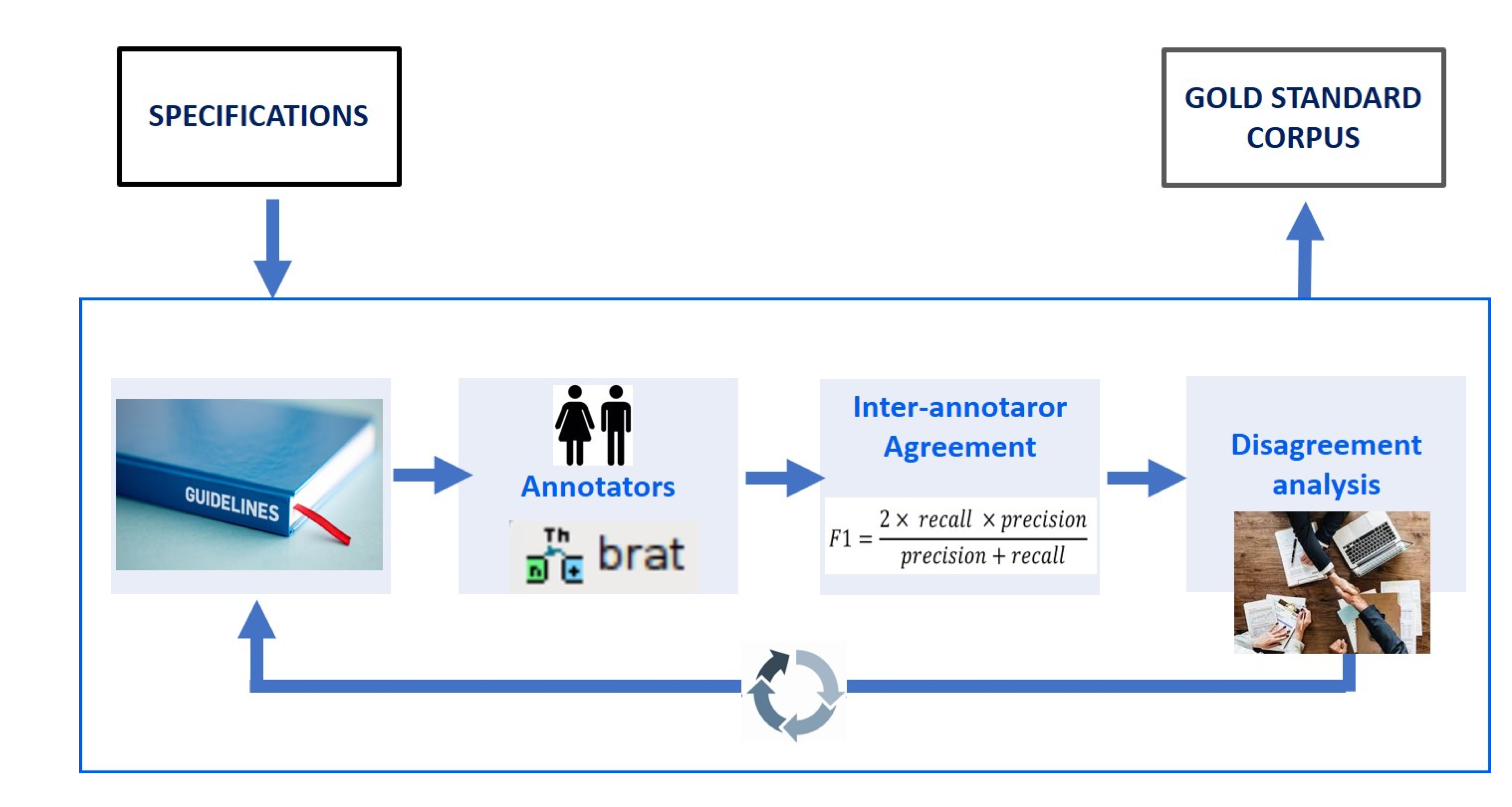}
     \caption{Annotation process of the RareDis corpus.}
      
\label{fig:Figure1}
      \end{figure}

After the automatic pre-annotation of the corpus, four people with experience in the creation of biomedical text corpora \citep{segura2010resolving,herrero2013ddi,krallinger2015chemdner,segura2015exploring,segura2017simplifying} and strong background in biomedicine and experimental dermatology of rare diseases \citep{chacon2019fibroblast,guerrero2019assessment,martinez2020combined} participated in the annotation of the corpus (Fig. \ref{fig:Figure1}).

In the first phase (named Specification), the annotator group reviewed a random set of 30 pre-annotated texts to define the set of entity and relation types and create a first version of the annotation guidelines. Once the annotation guidelines were defined, two experts on rare diseases thoroughly read them and were trained to use the BRAT annotation tool  \citep{stenetorp2012brat}.
Moreover, the following resources were used to help annotators during the annotation task:
\begin{itemize}
    \item NORD (National Organization for Rare Disorders) website  \citep{raredb} provides valuable information to clarify possible uncertainties during the annotation task. 

    \item ORPHANET contains information about rare diseases. Its´ aim is to improve diagnosis and treatment of rare diseases and facilitate access to the information on this topic.   
    \item GARD (Genetic and Rare Diseases Information Center) was created by the National Center for Advancing Translational Sciences (NCATS) and by the National Human Genome Research Institute (NHGRI) from the National Institutes of Health (NIH). GARD provides information about rare and genetic diseases \citep{gard}.
\end{itemize}

The corpus was divided in half and the two annotators separately conducted the annotation of the entities and relationships proposed in the annotation guidelines. They reviewed the automatic annotations, and then added, removed or modified them to complete and fix all the possible mistakes resulted by the automatic annotation process. During the pre-annotation process, neither the\emph{
sign} and \emph{anaphor} entity types nor relation types were automatically annotated. Thus, these entity types and all the relation types were annotated from scratch. 

To assess the quality of the corpus, find possible disagreements between annotators and avoid inconsistencies in the annotation guidelines, the Inter Annotator Agreement (IAA) was measured using F1-measure. This was calculated from a random sample of 51 texts, firstly for entities only. Then, the multidisciplinary team discussed all the disagreements and resolved them. As a result of this disagreement analysis, the annotation was redefined, after reaching a consensus in the annotation process. All the steps were redone one more time and all the texts were re-annotated by the two annotators using the improved guidelines. The IAA was measured using the same sample a second time for the entities and a first time for the relations. Disagreements were re-analysed to produce the final version of the guidelines, which were used to create the RareDis corpus. This corpus can be considered as a gold-standard corpus because it was manually annotated and its quality was proved by the IAA measurement between different annotators. Moreover, the annotation guidelines (see supplementary material) generated are supposed to be clear and will lead future experts in the annotation process of rare diseases as well as their signs and symptoms. 

\subsection{Annotation Guidelines}

To design the annotation guidelines, several annotation guidelines of different available corpora were reviewed \citep{dougan2014ncbi,herrero2013ddi}. During the specification phase of the annotation process (Fig. \ref{fig:Figure1}), the annotator group proposed the entity and relationship types to be annotated.  
The annotation guidelines were defined through an iterative process, which ensured the consistency and quality of the RareDis corpus. The guidelines provide clear and accurate descriptions of entities and their relations, as well as illustrative examples to help during the annotation task. Tables \ref{table:entities} and \ref{table:relations} provide the definitions and some examples of the entity and relation types included in the RareDis corpus. 

\begin{table}
\centering
      \begin{tabular}{lp{180pt}p{80pt}}
        \hline
        Entity type   & Definition  & Examples\\ \hline
        Disease & "An abnormal condition of a part, organ, or system of an organism resulting from various causes, such as infection, inflammation, environmental factors, or genetic defect, and characterised by an identifiable group of signs, symptoms, or both" \citep{otoole2003encyclopedia}. & \emph{cancer, alzheimer, cardiovascular disease}\\
        
        Rare disease & "Diseases which affect a small number of people compared to the general population and specific issues are raised in relation to their rarity. In Europe, a disease is considered to be rare when it affects less than 1 person per 2000" \citep{orphanet1}&\emph{acquired aplastic anemia, Fryns syndrome, giant cell myocarditis}\\
        
        
        Symptom & "A physical or mental problem that a person experiences that may indicate a disease or condition; cannot be seen and do not show up on medical tests" \citep{cancer}&\emph{ fatigue, dyspnea, pain}\\
        
        Sign&"Something found during a physical exam or from a laboratory test that shows that a person may have a condition or disease" \citep{cancer}&\emph{inflammation, rash, abnormal heart rate, hypothermia}\\
        
        Anaphor & Pronouns, words or nominal phrases that refer to a disease or a rare disease (which is the antecedent of the anaphor) & \emph{This disease, These diseases} (Fig. \ref{fig:Figure2})

        \\ \hline
      \end{tabular}
      \caption{Description of the entities annotated in the RareDis corpus.}
      \label{table:entities}
\end{table}

\begin{table}
\centering

      \begin{tabular}{lp{280pt}}
        \hline
        Relation type   & Definition \\ \hline
        
        produces &
        relation between any disease and a sign or a symptom produced by that disease (Fig. \ref{fig:Figure3}.a) \\

        increases risk of & relation between a disease and a disorder, in which the disease increases the likelihood of suffering from that disorder (Fig. \ref{fig:Figure3}.d) \\
        
        is a &
        relation between a given disease and its classification as a more general disease (Fig. \ref{fig:Figure3}.c) \\

        is acron &
        relation between an acronym and its full or expanded form
 (Fig. \ref{fig:Figure3}.c). \\
 
    is synon &
        relation between two different names designating the same disease  (Fig. \ref{fig:Figure3}.b)\\

anaphora &
        relation of an anaphor entity with its antecedent. The antecedent must be a disease or a rare disease  (Fig. \ref{fig:Figure2}). \\
        \\ \hline
      \end{tabular}
      \caption{Description of the relation types annotated in the RareDis corpus.}
      \label{table:relations}
\end{table}

\begin{figure}
  \centering

    \centerline{\includegraphics[scale=0.30]{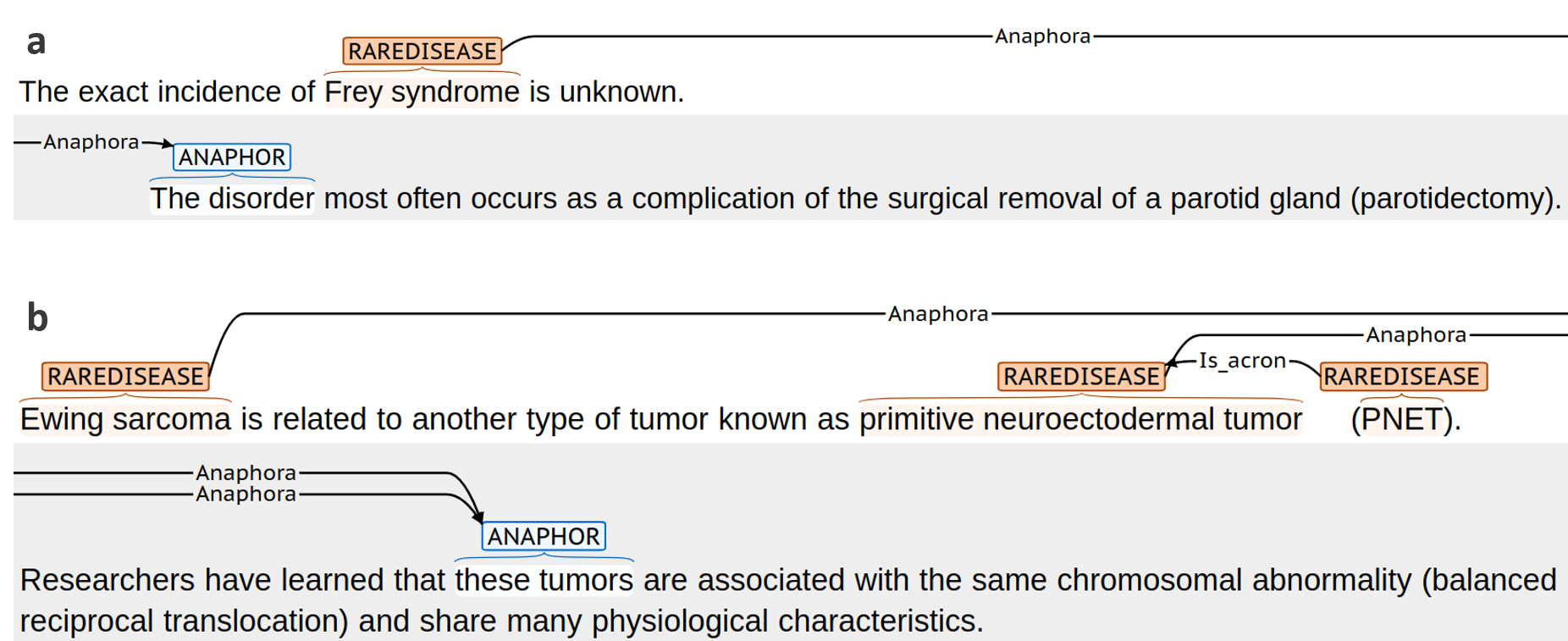}}
    \caption{Examples of anaphora annotations: 
     (a) Anaphor of one antecedent. (b) Anaphor of two antecedents }
\label{fig:Figure2}
      \end{figure}

\begin{figure}
  \centering

    \centerline{\includegraphics[scale=0.40]{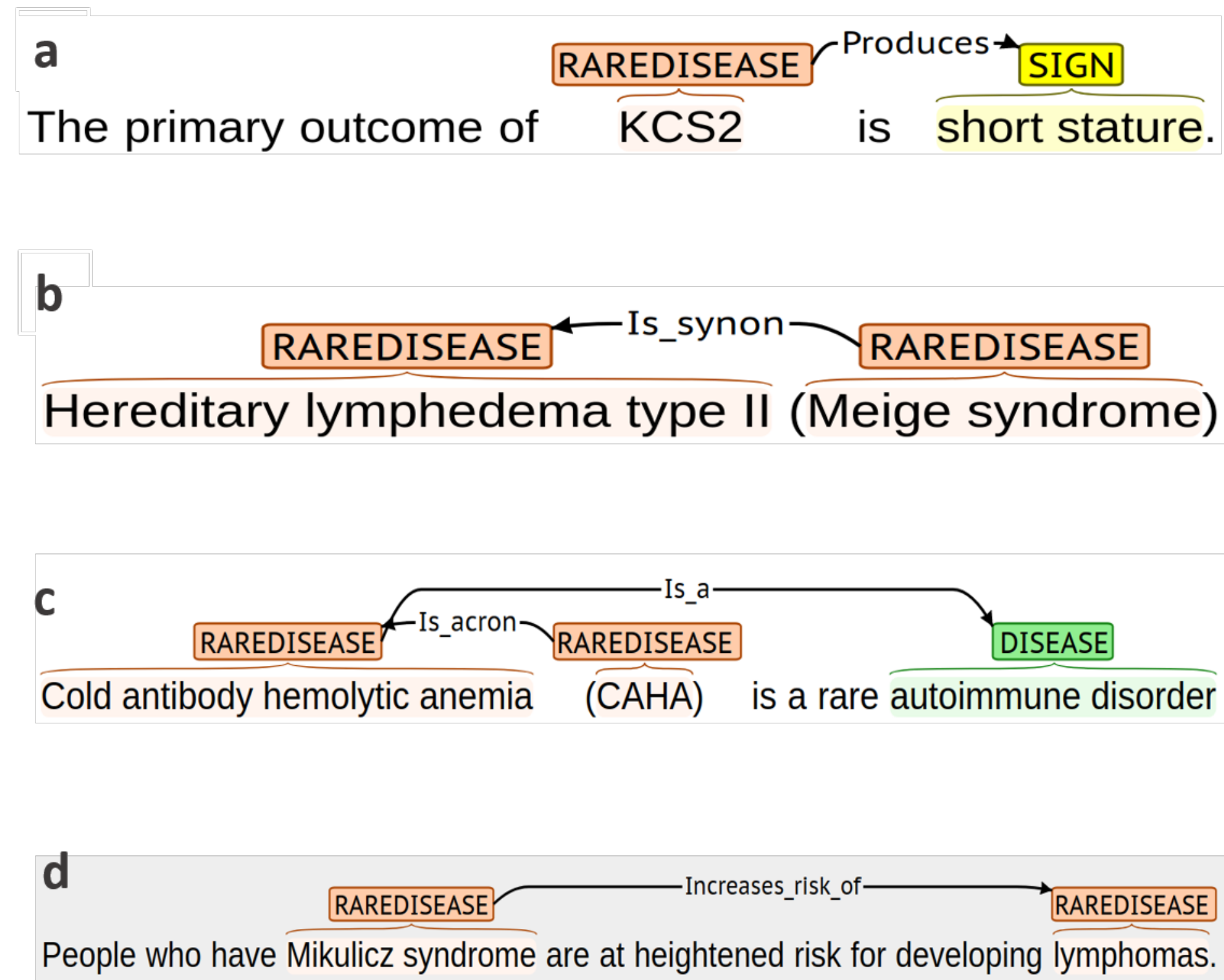}}
    \caption{Examples of the different relation types: 
      (a) Example of produces. (b)  Example of is synon. (c)  Example of is a and is acron. (d)  Example of increases risk of.}
\label{fig:Figure3}
\end{figure}
     
\begin{figure}
    \centering

    \centerline{\includegraphics[scale=0.30]{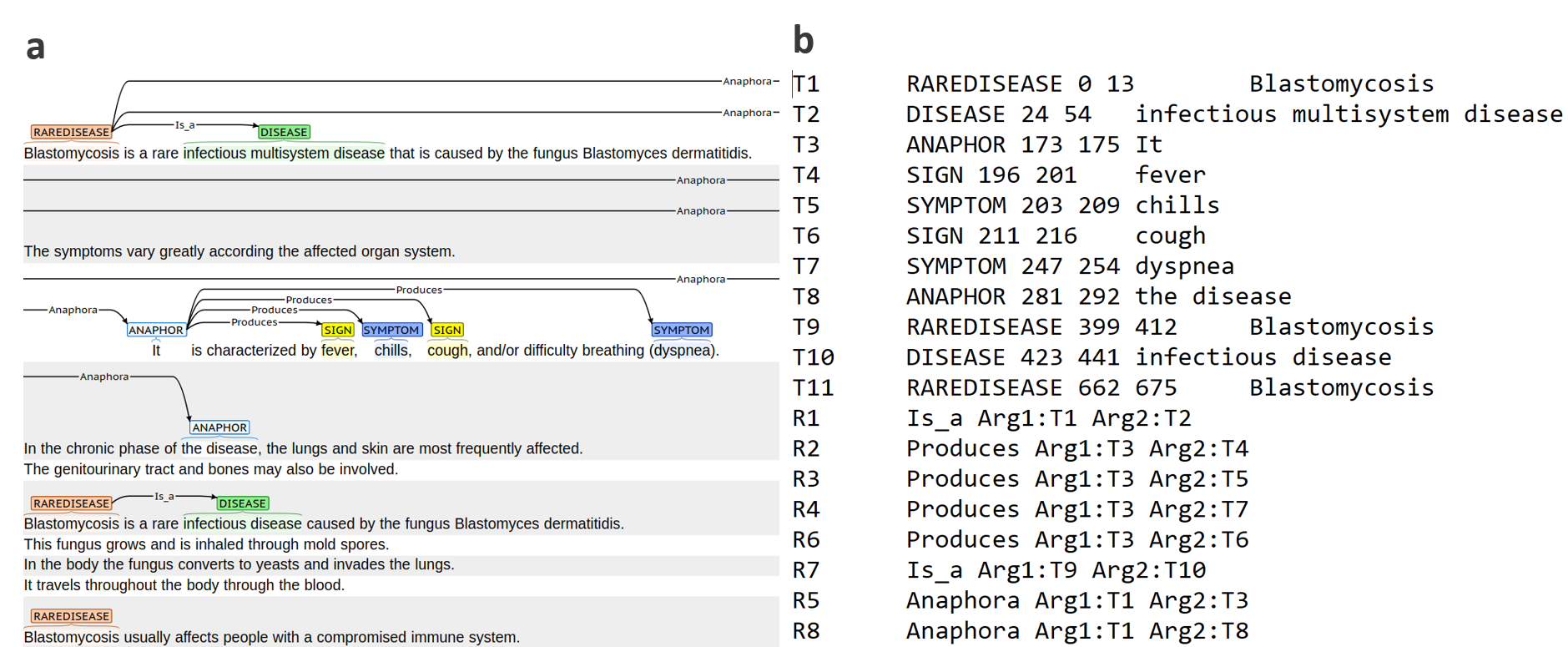}}
    \caption{Example of an annotated text with BRAT.}
\label{fig:Figure4}
\end{figure}

As previously mentioned, the BRAT annotation tool was used to perform the manual annotations. For each text file, this tool creates an ANN file containing the corresponding annotations for that text. This format has become a standard of corpora annotations for NLP tasks \citep{stenetorp2012brat}. Figure \ref{fig:Figure4} shows an example of text annotated with the BRAT tool and its annotation file with extension .ann, where the annotations are stored.

The final annotation guidelines for entities are summarised below.

\subsubsection{Anaphors and their relations}

Most of the previous studies on relation extraction  \citep{li2017neural,xing2020biorel,li2020bio,yamada-etal-2020-luke} are focused on the task at the sentence level, but extracting relations among entities in a paragraph is a more challenging task. In this regard, the RareDis corpus provides the annotation of anaphors. An anaphor is a linguistic unit referring to a previously mentioned linguistic unit in the text, which is named as an antecedent. Although we plan to annotate the anaphors referring to other entity types, the current version of the RareDis corpus only includes anaphoric expressions and their relations with their antecedents when they refer to mentions of diseases or rare diseases. 

  One of the main differences of the RareDis corpus compared to other corpora for relation extraction is the inclusion of relations annotations whose entities can occur in different sentences. The annotation of anaphors in the corpus could help to develop approaches capable of extracting relations described at the paragraph level. 

\subsubsection{Difference between signs and symptoms}

Signs and symptoms are abnormalities that may suggest a disease. However, they have different meanings. Signs can be detected by tests (e.g. \emph{low creatinine levels} or \emph{high blood pressure}) or observed by a physician (e.g. a \emph{erythema}, \emph{bleeding} or a \emph{lump}). On the other hand, symptoms are subjective indicators of a disease manifestation of disease noticed by the patient (for example, \emph{pain} or \emph{loss of appetite}). This difference should be considered during the annotation task. 

\subsubsection{Overlapped entities}

\begin{figure}
    \centering

    \centerline{\includegraphics[scale=0.40]{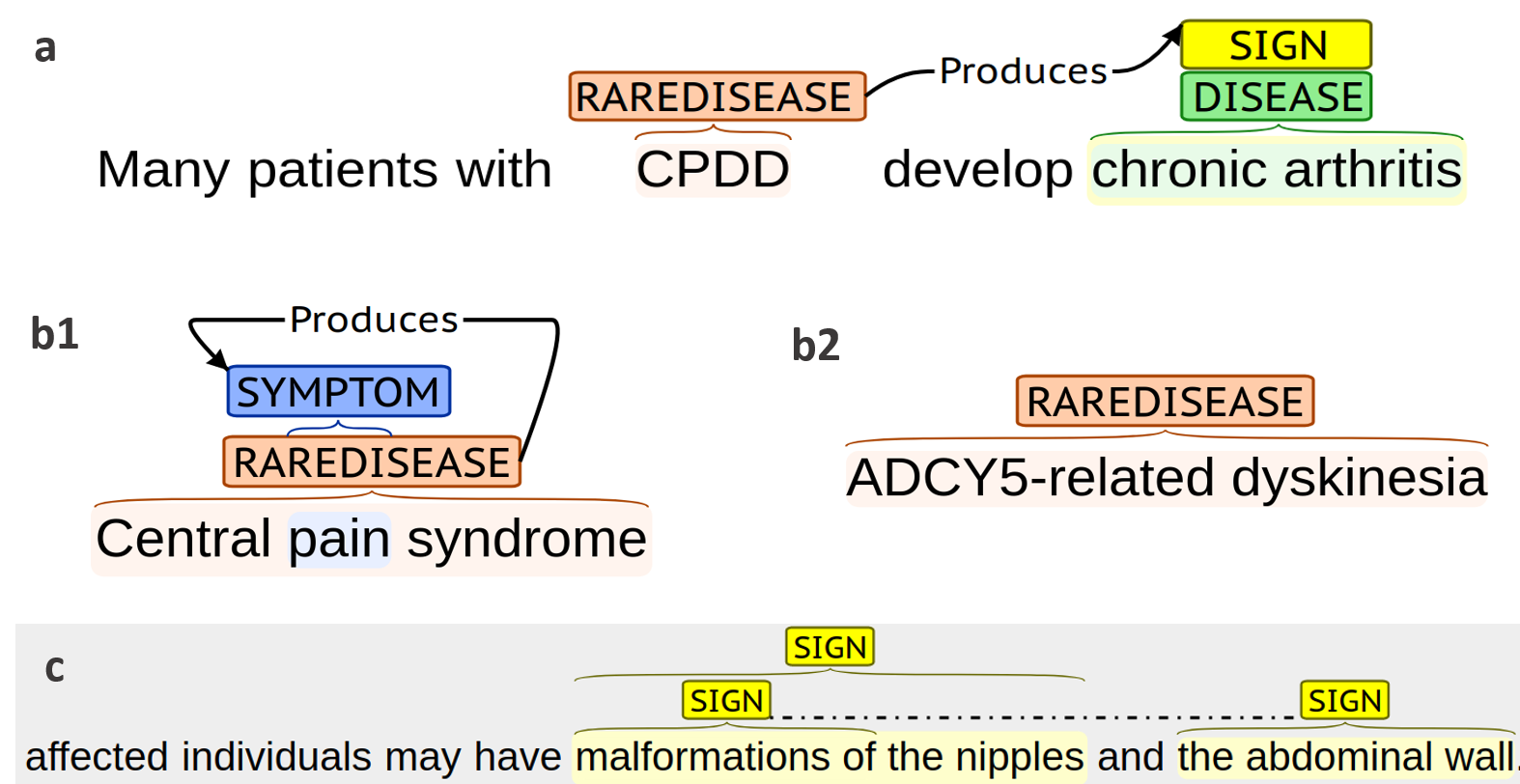}}
    
    \caption{Overlapped, nested and discontinuous entities examples:
      (a) Overlapped entity: Chronic
arthritis is annotated as a sign and as a disease. (b1)  Example in which nested entities have to be
annotated. (b2)  Example in which nested entities don´t have to be annotated. (c)  Example of a
discontinuous entity.}
\label{fig:Figure5}
\end{figure}

During the first iteration of the annotation process, some mentions were annotated with two or more different entity types. Figure \ref{fig:Figure5}.a shows an example where the same mention (\emph{Chronic arthritis}) was classified as a disease and also as a sign. 

We proposed the following rules when overlapped entities appear:
\begin{itemize}
    \item If a mention can be annotated as disease and rare disease, only the most specific one, i.e. rare disease, should be annotated. For example, \emph{acanthocheilonemosis} was automatically annotated as disease and rare disease, however, the most general type, i.e. disease,  was removed from the annotation.
    \item A disease can cause or be associated with other diseases (Fig. \ref{fig:Figure5}).a. In this case, the caused disorder should be annotated as disease, but also as a sign or symptom. Therefore, if a mention can be annotated as a disease and also as a symptom or sign, the annotators should include both entity types. 
\end{itemize}

\subsubsection{Nested entities}
Nested entities are mentions that are included in longer entity mentions. They are very common in texts from the biomedical domain (Fig. \ref{fig:Figure5} b1). Nowadays, the recognition of nested entities is still an unsolved problem. This is because most Named Entity Recognition (NER) systems developed so far are based on sequence labelling, in which each token can only be classified with a single label. This approach does not work for nested entities because their tokens may be classified with several labels. For this reason, most NER systems focus only on the recognition of the longest entity, without dealing with the inner nested entities.

In our corpus, the nested entities are also annotated if the inner and outer entities belong to different categories. For example, \emph{central pain syndrome}, which is a disease, also contains \emph{pain}, which is a symptom. In this case, both entities were annotated (Fig. \ref{fig:Figure5}.b1). 

However, there is an exception for this rule. When a nested entity (the inner mention) refers to a disease  and the longest mention is a more specific disease, only the most specific disease will be annotated. During the first iteration of the process annotation, this exception had not been defined yet. For this reason, the annotators annotated the most specific diseases (the longest mention) and also the more general disease (the inner mention). After the disagreement analysis and the revision of the annotation guidelines, the inner mentions of diseases were removed, keeping only the most specific mentions. In the example of Figure \ref{fig:Figure5}.b2, it can be observed how \emph{diskinesia}, which is a disease entity nested within \emph{ADCY5-related diskinesia}, is not annotated since it belongs to this exception.

\subsubsection{Discontinuous entities}

In addition to recognising nested entities, another major challenge in NER is how to deal with discontinuous entities. Most traditional NER systems make the assumption that an entity is a contiguous sequence of tokens (e.g. ADCY5-related dyskinesia). This is due to the fact that they are based on sequence labelling, which does not deal with the possible gaps in an entity mention. However, many entity mentions can be described as discontinuous sequences of tokens. These types of entities are even more frequent in the texts of our corpus, where they often describe the symptoms and signs of a disease. Each discontinuous mention should be annotated without including the tokens that do not belong to the mention (e.g. \emph{and} or the punctuation ','). In the example shown in Figure \ref{fig:Figure5}.c, it shows how to annotate the discontinuous entity \emph{malformations of the abdominal wall}.


\subsubsection{Abbreviations, acronyms and synonyms}
Abbreviations and acronyms are shortened forms of words or phrases. They are very common in biomedical texts, frequently used to refer diseases, drugs, and other biomedical entities. 
An acronym is usually composed of a set of initial letters of the tokens belonging to the entity (e.g. \emph{CAHA} is the acronym for \emph{Cold antibody hemolytic anemia}), while abbreviations are terms such as \emph{bid} (which means twice a day), \emph{Dr.} (which means Doctor), \emph{cap} (which means capsule) or \emph{post-op} (which means after surgery).

\begin{figure}
    \centering

    \centerline{\includegraphics[scale=0.40]{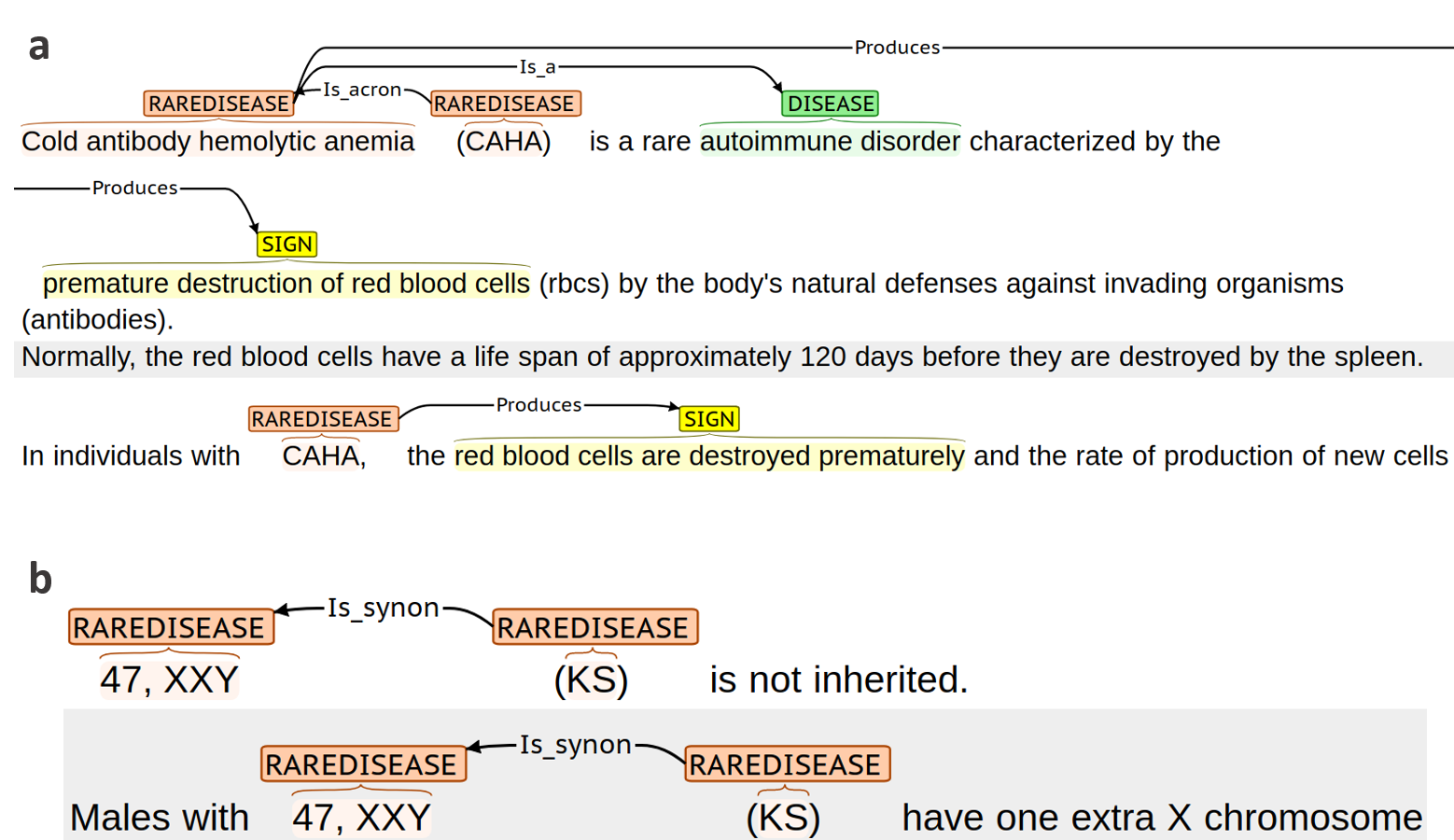}}
    \caption{Example of the annotation of \emph{is acron} and \emph{is synon}:
    (a) Example of annotation of \emph{is acron} once. (b) Example of a repeated annotation of \emph{is synon}.}
\label{fig:Figure8}
\end{figure}

The annotators should annotate these words when they refer to diseases or rare diseases. When an acronym and its long form occur in the texts, both mentions should be annotated with its corresponding entity type. Moreover, the annotators should also annotate the relation \emph{is acron} between both mentions. The same rules hold for the \emph{is synon} relation. Figure \ref{fig:Figure3}.c shows an example where the acronym \emph{CAHA} and its long form \emph{Cold antibody hemolytic anemia} were both annotated as rare diseases and related by using the relation \emph{is acron}.
However, this relation should only be annotated between the first mention of the acronym and its long form. If the acronym/synonym occurs more times in the same text, its following mentions should not be related to its long form or to its other name (see Figure \ref{fig:Figure8}.a), except when it is explicitly expressed in the text (see Figure \ref{fig:Figure8}.b).

Figure \ref{fig:Figure3}.b shows an example where a rare disease, \emph{Hereditary lymphedema type II} is followed by another mention of rare diseases, \emph{Meige syndrome}, between parenthesis. An automated method might misidentify \emph{Meige syndrome} as an acronym for \emph{Hereditary lymphedema type II}, when in fact it is not.
Therefore, in this case, the right relation between the long form and the mention between parentheses is \emph{is synon}. 

\subsubsection{Common names referring diseases, signs or symptoms}

\begin{figure}
    \centering

    \centerline{\includegraphics[scale=0.40]{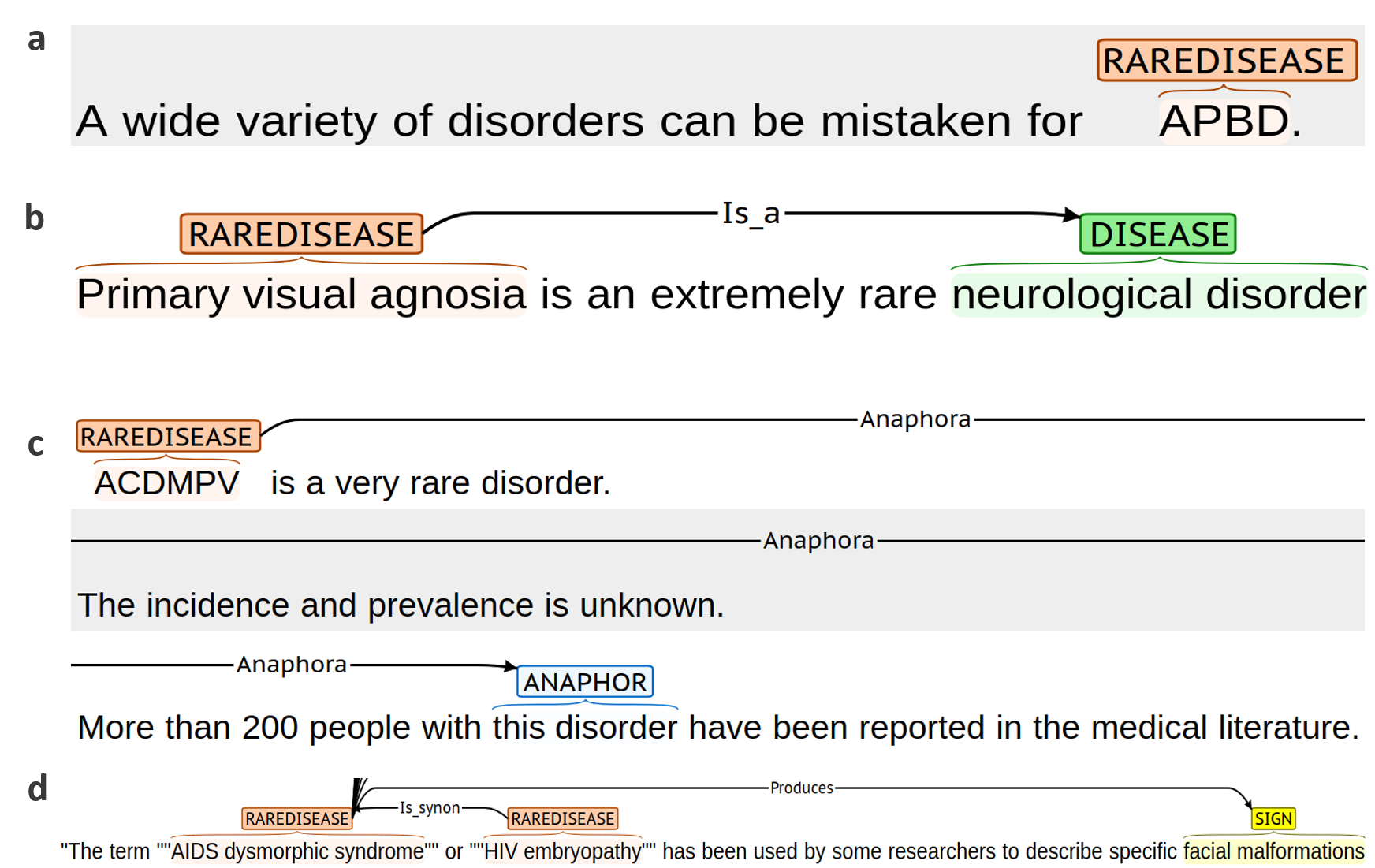}}
    \caption{Example of the annotation of general terms:
      (a) Example of a disease general term (\emph{disorders}) not annotated. (b) Example of general term of a disease with an adjective annotated (\emph{neurological disorder}). (c) Example of general term acting as anaphor. (d) Example of a sign general term annotated.}
\label{fig:Figure6}
\end{figure}

The selected texts contain numerous general terms that can refer to a disease, symptom or sign. These very general terms (e.g., condition, disorder, symptom or manifestation) should not be annotated (Fig. \ref{fig:Figure6}.a). 

An exception is when the general term is modified by an adjective. This was added as a clarification after the analysis of the disagreements.
In general, an adjective is used to describe or modify nouns in a sentence. Moreover, in our texts, adjectives preceding disease mentions allow us to refer to more specific diseases (such as  \emph{inherited disorder} or \emph{neurological disorder}). Although these mentions are general terms, we decided to annotate them with the entity type disease because they provide us information about other diseases that occur in the same paragraph. Figure \ref{fig:Figure6}.b shows an example where the phrase \emph{neurological disorder} was annotated, which allows us to know that \emph{primary visual agnosia} is a neurological disorder. On the other hand, a special case is made when the adjective does not define a subtype of a disease (e.g. \emph{symptomatic} or \emph{adult}).  

As with the names of diseases modified by adjectives, if an adjective (such as mild or severe, painful, etc.) modifies a symptom or sign, it should also be included in the annotation of the sign (e.g. \emph{mild diarrhea}, \emph{abnormal curvature of the spine}). The adjective \emph{asymptomatic} is considered a symptom.

Moreover, when the general term is within a nominal anaphor (e.g. \emph{the disease}, \emph{these disorders}) and its antecedent is a disease or a rare disease, the nominal anaphor should be annotated as \emph{anaphor} (Fig. \ref{fig:Figure6}.c), as explained before.

After the disagreement analysis, we defined new rules about the annotation of general terms such as \emph{abnormalities} or \emph{malformations}.  These should be annotated as signs because they could provide useful information about a disease (Fig. \ref{fig:Figure6}.d). 


\subsubsection{Annotation rules for  symptoms and signs entities}

Some signs or symptoms can be described by technical terms (e.g. \emph{proptosis}), or a lay description (e.g. \emph{protruding eyes}).
The disagreement analysis after the first iteration of the annotation task revealed that these cases were very challenging for the annotators. One of them annotated the two descriptions, while the other only annotated the technical term. After the discussion, it was decided that if both technical and lay descriptions are used to describe a sign or symptom, the annotators should only annotate the technical term, while the lay description should not be included in the annotation. For example, in the sentence \emph{paleness of the skin (pallor)}, the annotators should not annotate \emph{paleness of the skin}, since the technical name \emph{pallor} is present. Therefore, only \emph{pallor} should have been annotated as a sign.

If there is not a technical term after the description, all the description is annotated in the most concise way, making use of discontinuous annotations if necessary (Fig. \ref{fig:Figure5}.c).

Moreover, if the description of a sign or symptom also specifies the body part affected or the period of time during which the sign/symptom happens (e.g. \emph{prenatal}, \emph{during the childhood}), these should be included in the annotation. Some examples are: \emph{stiffness on one side of the face}, \emph{lesions in the gastrointestinal tract}, or \emph{postnatal growth retardation}.

The following sentence \emph{Affected individuals develop characteristic loss of body fat (adipose tissue)} contains the sign \emph{loss of body fat}. The term \emph{adipose tissue}, which appears within parenthesis, should not be included in the annotation of this sign, because this is a synonym of \emph{fat} and does not provide extra information of the sign.

\subsubsection{Annotation rules for relations}

Relations should be always annotated at the sentence level when both involved mentions occur in the same sentence (Fig. \ref{fig:Figure8}.a). If there are several mentions for the same entity in the text, but these belong to different sentences, the annotators should annotate as the first term of the relation, the one appearing just before in the text, except for  \emph{is synon} and  \emph{is acron} relations. Thus, in Figure \ref{fig:Figure8}.a, \emph{CAHA} appears two times in the same text. The text describes a relation between this rare disease and the sign \emph{red blood cells are destroyed prematurely}. This relation should not be included for the first mention of \emph{CAHA}, which occurs at the beginning of the text.

\begin{figure}
   \centering

    \centerline{\includegraphics[scale=0.50]{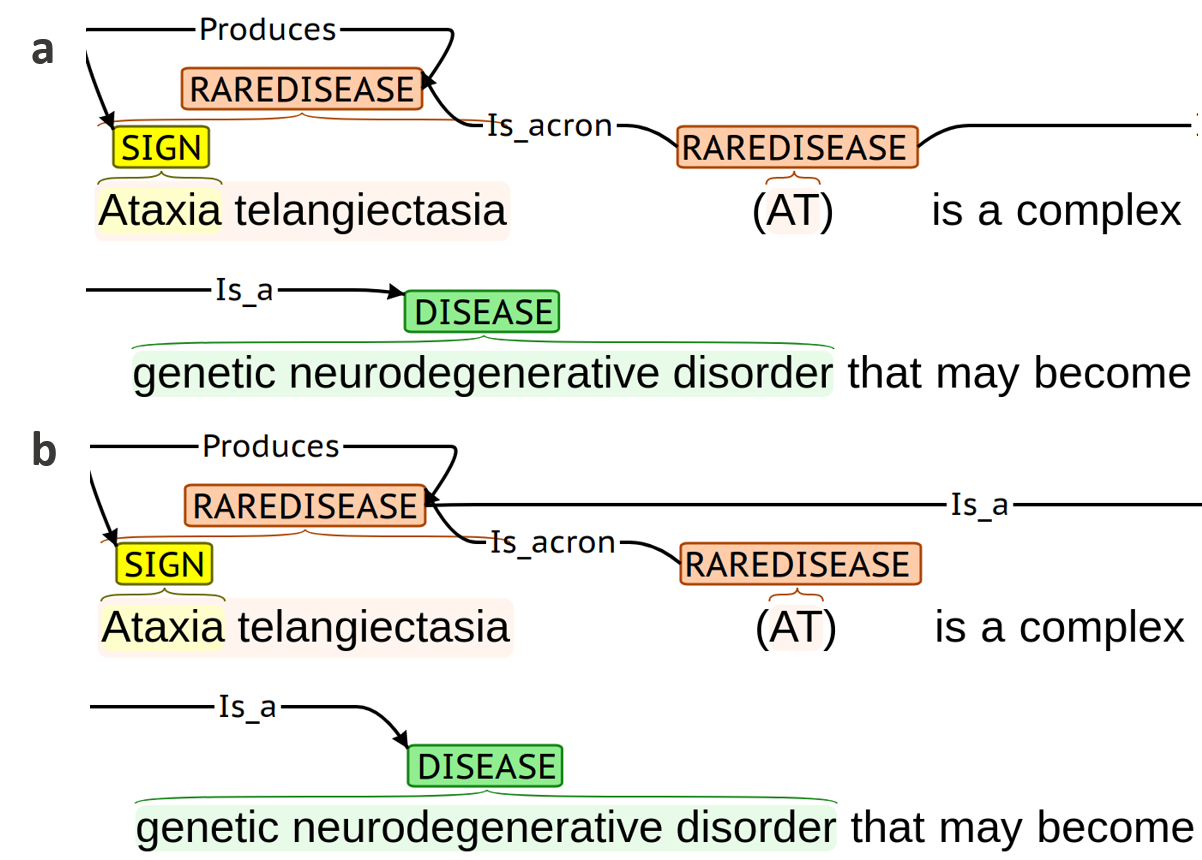}}
     \caption{Disagreement of a relation:
  Incorrect annotation (a)  and correct annotation (b)}
\label{fig:Figure7}
\end{figure}

In addition to this, we had to deal with relations where some of the involved entities appear two or more times in the same sentence (Fig. \ref{fig:Figure7}), as in the following sentence \emph{Ataxia telangiectasia (AT) is a complex genetic neurodegenerative disorder that may become apparent during infancy}, which describes an \emph{is a} relation between \emph{Ataxia telangiectasia}, a rare disease, and the disease entity \emph{genetic neurodegenerative disorder}. But, this sentence also contains an acronym, \emph{AT}, of \emph{ataxia telangiectasia}. At first, the first annotator only annotated the relation shown in Figure \ref{fig:Figure7}.b, while the second annotator included the relation between the acronym \emph{AT} and the sign. Therefore, in these kinds of sentences, we decided not to include the relation with the acronym. The same rule is applied for synonyms (Fig. \ref{fig:Figure3}.b).



\subsubsection{Associations, organisations, or organisms}
Many associations, organisations and organisms dedicated to specific rare diseases, contains or share the same name of the disease represented.  For example, ADNP is the acronym of a rare disease, and \emph{ADNP Kids Research Foundation} is the name of an organisation  to fund research for ADNP. When the disease mention is contained or refers to an association or organisation, it should not be annotated. 

\subsubsection{Genes}

Sometimes, genes and diseases share the same name. When the mention refers to a gene, this should not be annotated. 
For example, TRPS1 is the acronym of a rare disease, but in the sentence \emph{Molecular genetic testing can reveal mutations of the TRPS1 gene},  it refers to a gene, so it is not annotated.

The complete annotation guidelines are included in supplementary material.

\section{Results}
\label{sec:results}
\subsection{The RareDis corpus statistics}

The main goal of this work is to provide BioNLP community with an annotated corpus that can be used for learning and evaluating different machine learning models to extract valuable information about rare diseases and their clinical manifestations from texts. The annotations included in our corpus are examples that can be exploited by these algorithms to train models capable to detect similar information from unannotated texts. Therefore, once the corpus was annotated, we split it into training, validation, and test datasets in the ratio 70:10:20. The training dataset contains the annotated texts that will be used to train and learn the models. The validation dataset consists of the annotated texts that will be used to tune the parameters of each model, and the test dataset will be used only to evaluate the models.

Table \ref{fig:tokens} shows some basic statistics about the number of tokens, sentences and documents in the whole RareDis corpus, as well as, in its three subsets.

\begin{table}[ht]
\centering

      \begin{tabular}{ccccc}
        \hline
           & Training  &Validation   & Test & Total\\ \hline
        Documents & 729 & 104 & 208 & 1,041\\

        Sentences & 6,451 & 903  & 1,787 & 9,141\\
        Tokens & 135,656  & 18,492   & 37,893 & 192,041\\ \hline

      \end{tabular}
      \caption{Number of documents, sentences and tokens in the RareDis corpus.}
      \label{fig:tokens}
\end{table}

Table \ref{tab:raredis}  shows the numbers of the annotated entities and relations in the RareDis corpus. The frequencies of the entity type \emph{anaphor} and the relation \emph{anaphora}, as it was expected, are very close. The fact that the numbers do not exactly coincide is explained because some anaphors might refer to several diseases (Fig. \ref{fig:Figure2}), which results in a larger number of relations than entities.

\begin{table}[ht]
\centering

      \begin{tabular}{ccccc}
        \hline
           & Training  &Validation   & Test & Total\\ \hline
        Disease & 1,647 & 230 & 471 & 2,348\\
        Rare Disease & 3,608 & 525  & 1,088 & 5,221\\
        Symptom & 319  & 24   & 53 & 396\\ 
        Sign & 3,744  & 528   & 1,061 & 5,333\\ 
        Anaphor & 1,108  & 151   & 276 & 1,535\\ 

        produces & 4,106 & 556 & 1,131 & 5,793\\
        increases risk of & 169 & 22  & 54 & 245\\
        is a & 693  & 88   & 194 & 975\\ 
        is acron & 186  & 34   & 68 & 288\\  
        is synon & 80  & 16   & 15 & 111\\  
        anaphora & 1,113  & 151   & 279 & 1,543\\ \hline
        
      \end{tabular}
      \caption{Number of  entities and relations in the RareDis corpus.}
      \label{tab:raredis}
\end{table}





The most common entity and relation types are sign and produces, respectively. This may be because the main focus of the texts (which were collected from the NORD database) is  the clinical manifestations of rare diseases. The second most common entity type is rare disease, since every text of the corpus describes at least one rare disorder, whose name usually appears several times within the same text. All the entity and relation types show similar distribution in the three datasets. 

\begin{figure}
    \centering

    \centerline{\includegraphics[scale=0.70]{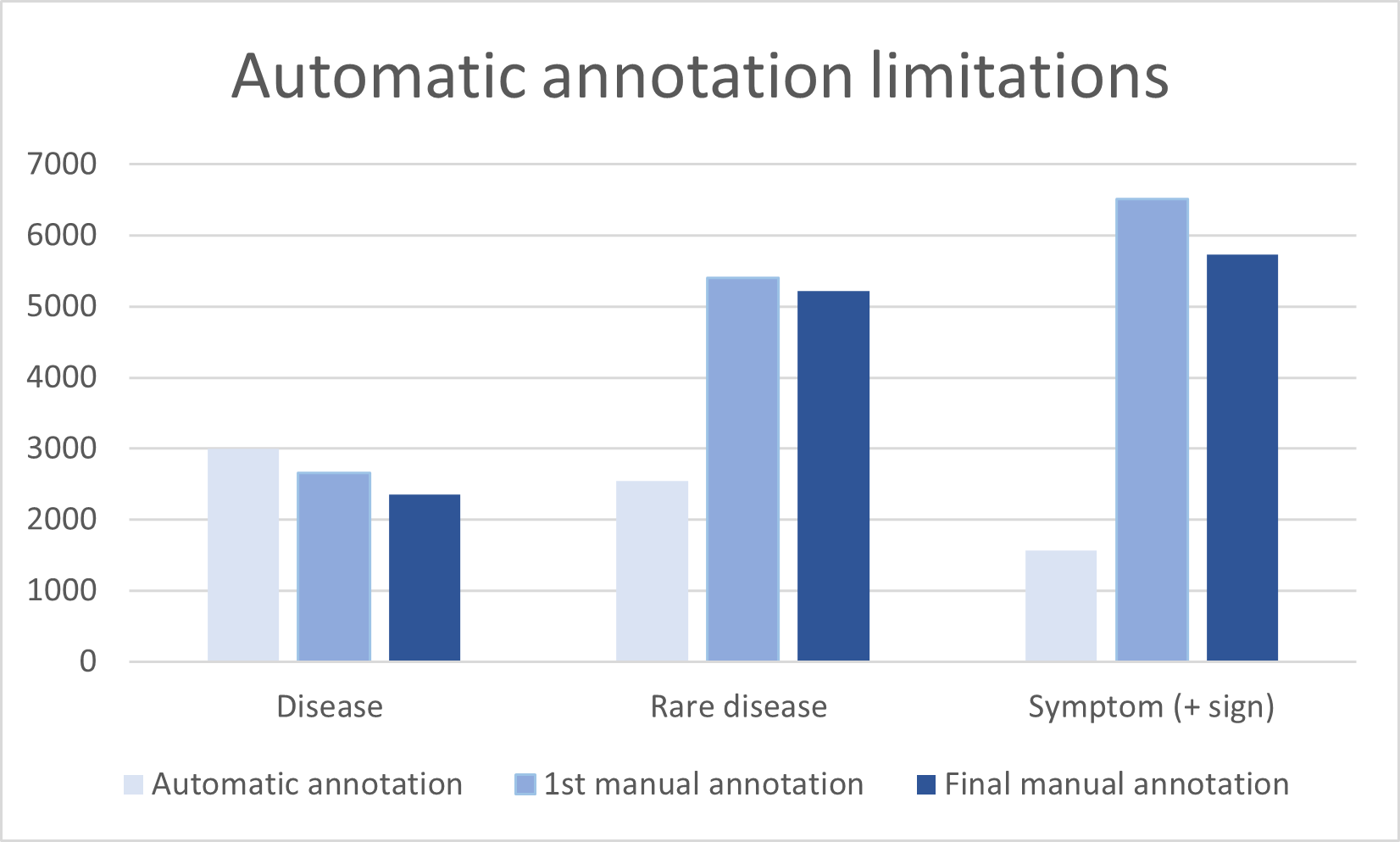}}
    
    \caption{Comparison between automatic and manual annotations.}
\label{fig:Figure9}
\end{figure}

All annotations for anaphors and relations were done from scratch without using dictionaries or automatic tools. The Symptom Ontology (SYMP) did not distinguish between symptoms and signs, and thereby, all mentions detected using this resource were initially annotated as symptoms, and later, the annotators manually corrected the mentions that referred to signs. We have studied the differences between the automatic and manual annotation for diseases, rare diseases, symptoms and signs. It can be observed from Figure \ref{fig:Figure9} that the lowest variation is for diseases. This could be explained because the terminological resources for diseases are often more comprehensive than for other entities such as symptoms or signs. Regarding the difference of the automatic and manual annotations for rare diseases, the possible causes may be: 1) some rare diseases had already been detected as common diseases using the dictionaries, 2) the dictionaries do not contain most of the acronyms for rare diseases, and 3) some rare diseases were not included in the dictionaries. 
The greatest variation is found for symptoms and signs. The dictionaries for these entity types have much less coverage than the dictionaries for diseases, as it was described in subsection \ref{subsec:construction}. Moreover, many mentions of signs are described by  short phrases or sentences, instead of using medical terms. Many of these phrases usually contain other issues such as nested entities, overlapped entities or discontinuous entities, which can not be accurately addressed by the dictionaries.  

\subsection{Inter-Annotator Agreement}

The inter-annotator agreement (IAA) not only assesses the consistency and quality of the corpus, but also establishes an upper threshold for results in the tasks of information extraction from texts about rare diseases. 

Although Cohen's Kappa is considered as the standard measure for IAA \cite{mchugh2012interrater}, it is not recommended for the NER task  \cite{grouinalpha2011proposal,hripcsak2005agreement}. The Kappa measure requires a number of negative cases. We could calculate this number on the token level, however, since the frequency of the O label is much higher than other labels, the Kappa score would be misguidedly too high. Another approximation could be to measure the Kappa score on those tokens that are part of some annotation, but this would yield a low Kappa score. Thus, the F1-measure has become the standard metric of IAA for NER \cite{deleger2012building}. Specifically, we considered the annotations made by the first annotator our gold standard. Then we calculated the precision, recall, and F1-measure for the annotations created by the second annotator. In both cases, entities and relations, the F1-measure was calculated by checking the consistency under exact match criteria of both annotators.  Thus, the entity annotations should exactly coincide by not only the entity type assigned to a given mention, but also by coinciding exactly in the mentions. In the case of relations, an exact match is achieved when both annotators choose the same pair of entities and the same relation type to represent a relation instance. To compute the F1-measure, the bratiaa library, which allows the measurement of the IAA for entities annotated with the brat format, was used \cite{bratia}. This library already allows to obtain the agreement under exact match (type and mention) for entities. For this task, we implemented an extension of this library (https://github.com/isegura/NLP4RARE-CM-UC3M) to calculate the agreement under exact matches (entities and relation type) for relations.

\begin{figure}
    \centering

    \centerline{\includegraphics[scale=0.60]{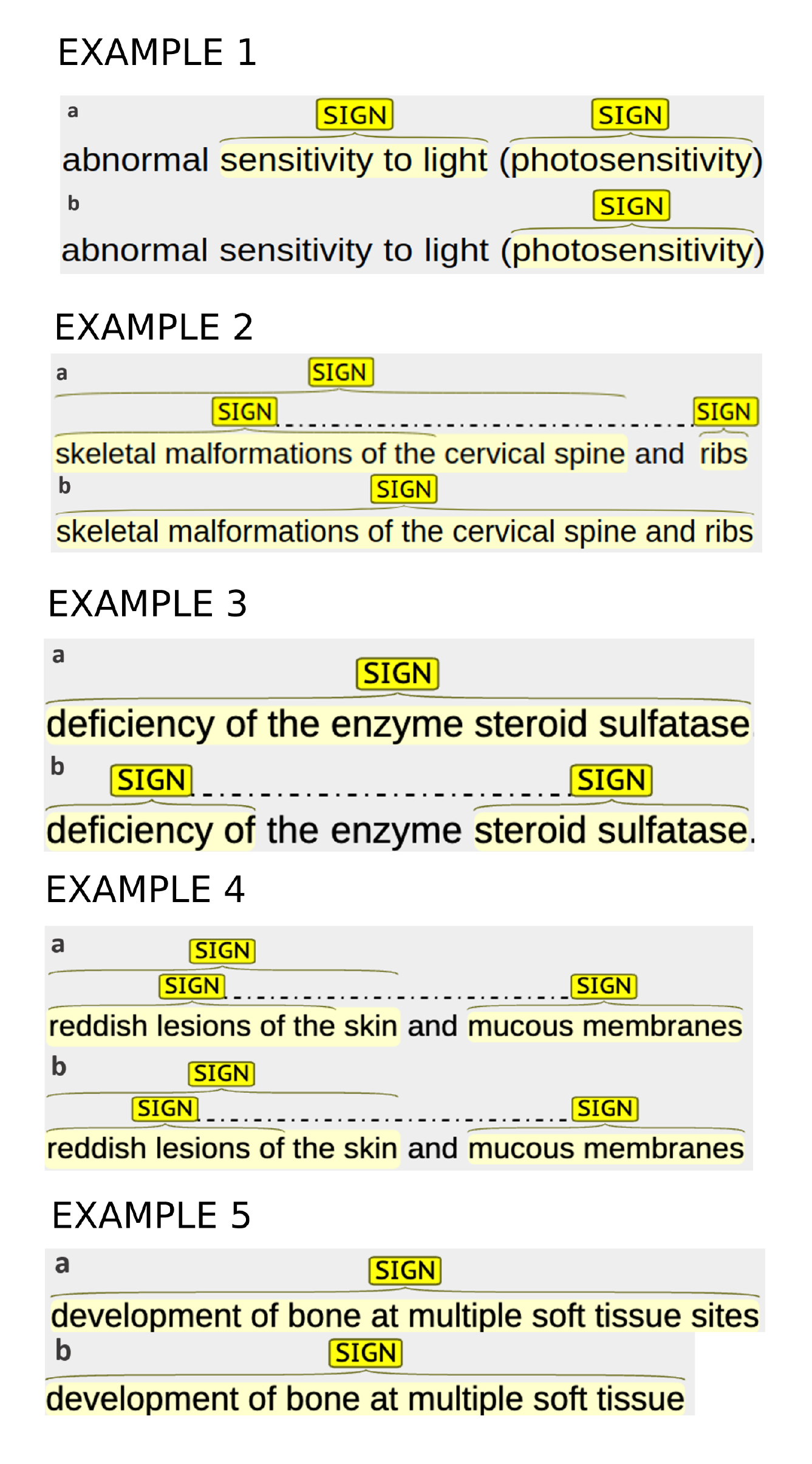}}
    \caption{Examples of disagreements for signs}
\label{fig:Figure10}
\end{figure}

During the annotation process two iterations were performed. After the first one the IAA was calculated obtaining an initial average of 62.6\%. Specifically, regarding sings, the first IAA was 48\%. After that, text were reviewed manually and ambiguous cases and disagreeements were discussed to redefine and clarify the annotation guidelines. The main causes of disagreements were:
\begin{itemize}
    \item Signs/symptoms descriptions: The first version of the guidelines proposed that if the description of a sign/symptom was concise and short, it should always be annotated, even if it appears together with its technical term. During this task, multiple disagreements between annotators were found in the criteria for defining when a description was concise and short. The first example of Figure \ref{fig:Figure10} shows one of these disagreements. Therefore, we decided to change this rule and annotated only the technical term.  If the technical term does not appear in the text, then, the annotators should annotate the description of the sign/symptom. Although this rule applies to both sings and symptoms, the vast majority of the cases actually refer to signs. 
    \item Nested entities are very common in the disease and rare disease entity types. They were also a common cause of disagreements between annotators. While the first annotator only annotated the longest mention, the second annotator also annotated the nested entities inside the longest one. As was explained previously in the guidelines, we defined some rules to deal with nested entities.

    \item	Discontinuous entities: Some mentions, especially those of signs, usually require the annotation of discontinuous entities, which may cause some disagreements in the detection of the gaps or even in the inclusion or not of some words in each mention. For example, the second annotator did not annotate many of the signs expressed by using coordinate structures (Fig. \ref{fig:Figure5}.c).  To avoid a low IAA in these cases, we included more examples in the guidelines to support the task of discontinuous entities annotation. Some additional examples are shown in Figure \ref{fig:Figure10}.
   
    \item Common names for diseases: it was difficult to differentiate between general terms of diseases that do not have to be annotated (e.g., disorder, condition or disease) and general diseases names (e.g., progressive disease or genetic disorder) that have to be annotated. To solve the disagreement arising from this, it was decided that if the general term (e.g., disorder or disease) appears together with an adjective (e.g., inherited, dominant, progressive or neurological), which is providing information about the disease, this mention together with the adjective should be annotated (Fig. \ref{fig:Figure6}.b). However, if disorder or disease appears alone, it should not be annotated.

\item Another common disagreement came up when annotating very general signs (such as abnormalities or malformations), because the second annotator considered them as too general, and therefore, did not annotate them. To solve them, a clarification was added to the guidelines remarking that these types of cases should be annotated.

\end{itemize}

\begin{figure}
    \centering

    \centerline{\includegraphics[scale=0.47]{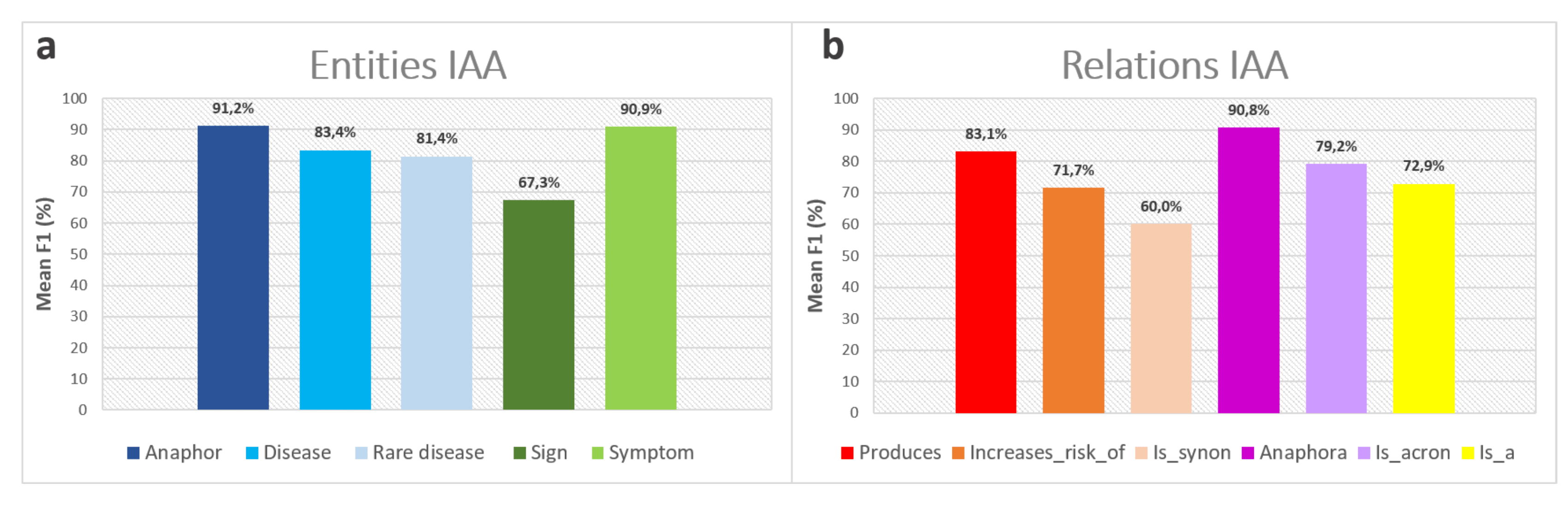}}
    \caption{IAA scores for entities and relations: 
      (a) IAA scores for entities. (b) IAA scores for relations}
\label{fig:FigureIAA}
\end{figure}

The final IAA value for entity types is 83.5\%. This is considered a very high F1-measure, representing a substantial agreement between annotators. This corroborates that the annotation guidelines are quite clear and specific and that they can be used as standard annotation guidelines for rare diseases, their signs and symptoms. These high-quality guidelines made possible the creation of a high-quality corpus. 

In order to calculate the IAA of the relations, the entities were manually reviewed according to the final guidelines. That is, both annotators annotated the relations based on the same set of entities. The IAA value for relation types is 81.3\%. Figure \ref{fig:FigureIAA}.b shows the IAA values per relation type.

The most common cause of disagreement for relations was the presence of sentences describing a relation where some of the involved entities appear two or more times in the text. There was no consistency in the annotation of these relations. To avoid this, in the guidelines it is stated that for relations annotated among different sentences, the first entity of the relation should be the one appearing first in the texts, except for the \emph{is synon} and \emph{is acron} relation types. 


\section{Discussion}

We present herein the generation of the RareDis corpus, that includes the annotation of rare diseases and their clinical manifestations (symptoms and signs). The annotation scheme followed (Figure \ref{fig:Figure1}) and the guidelines (see Supplementary Material) have been defined based on its utility to researchers and clinicians in the domain of rare diseases. 

The RareDis corpus has a similar size, in terms of documents and disease mentions, to previous corpora annotated with diseases  \cite{dougan2014ncbi}. Its´ size is enough to train and test supervised machine learning approaches for recognising \emph{diseases, rare diseases} and \emph{signs}, as well as the \emph{produces} relation. Moreover, the annotation of anaphoric expressions and their relations with their antecedents also results in a valuable resource for training systems for anaphora resolution in the biomedical domain. The annotation of signs could also address some NER challenges, such as nested entities or discontinuous entities, that have hardly been addressed. However, the number of instances for some entity types (e.g. \emph{symptom}) and most of the relation types are too small for training supervised machine learning approaches. For these types, the RareDis corpus could be used to develop semi-supervised approaches where it provides gold-standard seeds to augment the training data. 

In our analysis of the RareDis generated corpus, the measure of the IAA score, was performed to ensure its quality and consistency, which allows the complexity of the annotation task to be determined as well as providing insights into the quality of the guidelines developed. Furthermore, IAA values also provides an upper threshold for NLP systems that can extract useful information about rare diseases. IAA values indicate the high quality of the RareDis corpus (see Fig. \ref{fig:FigureIAA}). IAA results show very high agreement for anaphor (91.2\%), symptom (90.9\%), disease (83.4\%) and rare disease (81.4\%) entities compared to moderate agreement for signs (67.3\%) (see Fig. \ref{fig:FigureIAA}.a).  
The lower IAA for signs may be due to the intrinsic particularities of the annotation of this entity type, considering that many of them are discontinuous entities or are described by a short phrase instead of just by one or two technical words. Moreover, since many of them are not predefined, and for that reason, the annotator subjectivity could influence how these complex entity spans are annotated. On the other hand,  other entity types (diseases, rare diseases and symptoms) are usually composed of concrete predefined words, facilitating their objective identification and decreasing the disagreements associated to them. 

Figure \ref{fig:Figure10} shows additional examples of the main disagreements for signs. The fact that signs are primarily described by phrases, instead by just one or two words, already involves several problems related to the subjectivity of each annotator when including or not certain words within the annotation. One of the most important challenge is the accurate annotation of discontinuous entities. Different annotators might produce very similar annotations but with some small differences, as can be seen in Figure \ref{fig:Figure10}. In the second example, while the first annotator (a) correctly identified two different signs: ''skeletal malformations of the cervical spine" and ''skeletal malformations of the ribs", the second annotator wrongly considered both signs as a single one. In the third example, both annotators identified a single sign, however, they disagreed when including the phrase ''the enzyme". 
The fourth example shows an even more subtle disagreement than the previous ones. This example includes two different entities: i) ''reddish lesions of the skin" and ii) ''reddish lesions of the mucous membranes", which is a discontinuous entity. The first annotator (a) correctly identified them, however, the second annotator (b)
forgot to include the article ''the"  in the annotation of this discontinuous entity. Although both annotators  correctly detected the existence of a sign in this text, these small disagreements strongly penalized the global IAA for signs. 
The fifth example does not include any discontinuous entity, however, it very similar to the third example. Both annotators have practically identified the same entity, with the only difference that the second annotator decided not to include the word ''sites". Despite the existence of small disagreements between the annotators, we could consider that many signs have been correctly annotated and the information is not lost. Therefore, these partial disagreements are strongly penalizing the global IAA for signs, however, 
the same disagreements would be expressing semantically compatible mentions of signs.

IAA results also show very high agreement for relations. The anaphora type has the highest IAA (90.8\%), followed by produces with an IAA of 83.1\%. The annotation of the \emph{anaphor} entity and its antecedent is usually straight-forward, leaving less room for the subjectivity of the annotators, and thereby, for the disagreement. 
The \emph{is synon} relation type achieves the lowest agreement between the annotators (60\%), which can be considered a moderate agreement. This relation type only represents the  1.24\% of all the relation instances in the corpus. In fact, there are only five of such relation instances in the sample used to measure the IAA, so that, a single one disagreement may cause a very low IAA for this relation. 
The \emph{increases risk of} relation type obtains the second worst result, which might be explained due to the high dependence of the identification of this type on the context and on the annotator interpretation.

\section{Conclusions}

Thanks to biomedical research, a great deal of knowledge about rare diseases has been generated in recent years. The high cost of molecular analyses and the existence of a limited bibliography, sometimes inaccessible or scattered, hinders progress in the diagnosis and treatment of these conditions. Often affected patients, despite their fatal or chronically disabling conditions, lack an appropriate treatment. Rare disease diagnosis requires a high degree of expertise and specialisation. The identification of symptoms and clinical manifestations are essential steps in terms of making easier the diagnosis for clinicians. 

The RareDis corpus can serve as a gold standard for the development of NLP approaches for increasing the knowledge about rare diseases. As future work, we plan to extend our corpus to include scientific articles, clinical notes and  clinical cases about rare diseases. The extended corpus could potentially help to achieve a significant improvement in diagnosis velocity and treatment choice for patients suffering  rare diseases. For example, the \emph{produces} relation could be useful in improving patients’ diagnoses and  \emph{increases risk of}  to prevent the further development of complex phenotypes and other complications.  The importance of time in the first years of the manifestation of rare diseases  has been demonstrated as a key issue in the prognosis of patients and their quality of life.  \cite{brooks1999earlier} In the context of treatment also, recognising the disease in a more accurate way it is crucial for the personalised medicine of these patients.  \cite{baxby2014early} 

Information extraction techniques developed using the extended corpus could structure the information about rare diseases in smarter and more efficient way, solving the  difficulty of dispersed information among different sources. 
In addition to this, and important as well, these techniques can find relations between different diseases. This could facilitate, for example, the process of drug repositioning, which is a usual therapeutic strategy for orphan drugs and rare diseases \cite{jourdan2020drug,scherman2020drug,xue2018review,govindaraj2018large,sardana2011drug}. Thus, finding shared clinical manifestations among diseases could help to understand better the diseases’ mechanisms and to purpose new treatment strategies. 


\section*{Author's contributions}

ISB and SGA conceived and designed the research, conducted the literature search, methodology, writing-reviewing, and supervision. They also collaborated to define the guidelines and solve the disagreements.
CMM collaborated to define the annotation schema, process, and  guidelines. She was one of the annotators and analysed the main causes of disagreements. She was also responsible for the final version of the annotations as well as the guidelines. CMM, SGA and ISB wrote the paper. 
ECS helped to define the guidelines and solve disagreements. He was the second annotator. CMM and SGA  elaborated the figures, tables, and graphs.
All authors read and approved the final manuscript.

\section*{Funding}
This work was supported by the Madrid Government (Comunidad de Madrid) under the Multiannual Agreement with UC3M in the line of "Fostering Young Doctors Research" (NLP4RARE-CM-UC3M) and in the context of the V PRICIT (Regional Programme of Research and Technological Innovation; the Multiannual Agreement with UC3M in the line of "Excellence of University Professors (EPUC3M17)"; and a grant from Spanish Ministry of Economy and Competitiveness (SAF2017-86810-R).

\section*{Acknowledgements}
Authors are indebted to Ellen Valentine for grammar and stylistic corrections.




\end{document}